\pdfoutput=1
\documentclass[11pt]{article}

\usepackage{ACL2023}
\usepackage{booktabs} 
\usepackage{times}
\usepackage{latexsym}

\usepackage[T1]{fontenc}
\usepackage[utf8]{inputenc}

\usepackage{microtype}

\usepackage{inconsolata}

\usepackage{graphicx}
\usepackage{xcolor}
\newcommand*{\rom}[1]{\expandafter\romannumeral #1}
\newcommand{\ie}{\textit{i}.\textit{e}.}

\title{Sample Efficient Multimodal Semantic Augmentation for \\Incremental Summarization}
\author{ \\
    \\
  Intel labs,USA
  }
\author{ \\
  Intel labs,USA
  }
\author{Sumanta Bhattacharyya\thanks{ \hspace{0.1cm} Work done while at Intel labs} \\ UIC, USA 
        \And
        Ramesh Manuvinakurike \\ Intel labs, USA 
        \AND
        Sahisnu Mazumder \\ Intel labs, USA 
        \And
        Saurav Sahay \\ Intel labs, USA 
        }

\begin{document}
\maketitle
\begin{abstract}
%Semantic augmentation plays an important role in summarization. These semantic concepts along with input can help pre-trained models to learn a better understanding of the correlation between the semantics and input hence enriching the summary. We propose a cascaded model where we extract these semantic concepts  as an intermediate step using a pre-trained vision-language model (i.e.CLIP) and use these concepts along with the summarizer model's input to get the summarizer output. In order to improve the semantic augmentation we have modified the CLIP model's architecture inspired by recent Flamingo work also in order to improve the learning we introduced a clustering-based batch selection approach to ensure diversity in each batch while preserving the temporal relationship between the frames describing the same annotation. Our work also explained which portion the frames sampled from a frame pool describing an event have the most impact in extracting these augmentations. We have experimented on YouCook II dataset. Our method improves the summarization in terms of both Rouge Recall and precision scores ( around 10\% increase in terms of Rouge precision score and 8\% increase in terms of Rouge recall score from the base pretrained model). We have used Distilbart and BART-base for base pretrained model as summarizer. 

In this work, we develop a prompting approach for incremental summarization of task videos. We develop a sample-efficient few-shot approach for extracting semantic concepts as an intermediate step. We leverage an existing model for extracting the concepts from the images and extend it to videos and introduce a clustering and querying approach for sample efficiency, motivated by the recent advances in perceiver-based architectures. 
Our work provides further evidence that an approach with richer input context with relevant entities and actions from the videos and using these as prompts could enhance the summaries generated by the model. 
We show the results on a relevant dataset and discuss possible directions for the work. 

\end{abstract}

\section{Introduction}
Summarization is the consolidated format for a large document and has been widely used for many applications \ie, understanding a long meeting/event, story summarization etc. Abstractive summarization is challenging in the Natural Language Generation(NLG) domain as it requires an understanding of all the salient information in the input document and rewriting logically in a condensed manner rather than selection (extractive). Recent advancements in transformer-based abstractive summarization have shown promising attempts \cite{su2020two,hoang2019efficient,wang2020friendly} with ideas ranging from  the two-stage method,domain-adaptive training to plug and play topic models on top of the transformer. Despite these strong advancements in text-based summarization, there is a huge potential for how we can improve summarization from multimodal data. Since in real-time, data prevails in different modes rather than a single mode like text, there has been an increasing demand for how we can bridge the gap between these modalities \ie, cross-modal search applications for video, utilize the text data associated with the video to search for relevant video content \cite{
otani2016learning,song2011multiple}, which requires a complete understanding of the video without ignoring the subtle differences \cite{wang2012event}. Recent work \cite{palaskar2021multimodal} suggests that learning a semantic concept as an intermediate step can help the model to learn efficiently. Learning a semantic concept has always been beneficial in categorization tasks like scene recognition, video tagging, etc \cite {zhou2017places,ghadiyaram2019large}. 

\par
Recent advancements in the vision-language-based models \cite{radford2021learning,alayrac2022flamingo} have shown immense potential for generating text-based descriptions from images/videos. In our context, we refer to these text-based descriptions as "semantic concepts". Our work utilizes learning of these semantic concepts as an intermediate step from the videos. These semantic concepts along with the transcriptions (semantic augmentation) as input to a pre-trained summarizer model enrich the performance. In this work, we address the problem of (\rom{1}) generating semantically relevant annotations of a video (semantic concepts) using a fixed number of sampled frames from each video segment. (\rom{2}) utilize these semantic concepts along with input transcription (semantic augmentation) to enrich the summarization output of pre-trained models. 
(\ie BART).

In summary, Our contributions are the following:
\begin{itemize}
    
    \item We propose a novel CLIP-based approach \cite{radford2021learning} to generate semantic concepts from video frames. 
    \item In order to maintain diversity in each batch, we propose a clustering-based batch creation approach.
    \item We have experimented with our proposed approach using the YOUCOOK2  \cite{ZhXuCoAAAI18} dataset. The results perfectly demonstrate the efficiency of our approach.

\end{itemize}{
}

\begin{figure*}[h!]
  \centering
  \includegraphics[width=\linewidth]{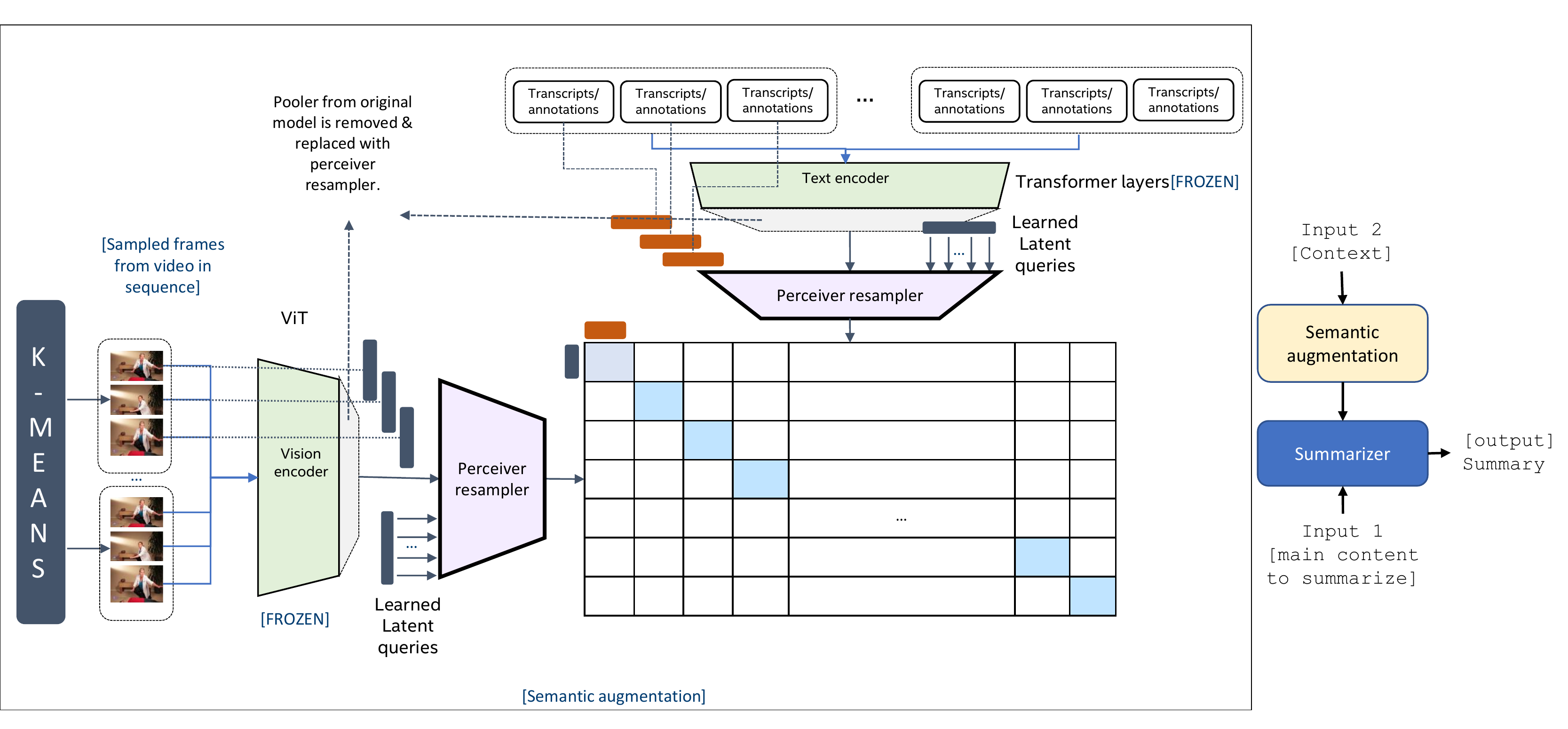}
    \caption{Shows the architecture of the system and the use of semantic augmentation for summarization.}
\end{figure*}

\par

\section{Related work}
Early attempts show promising ideas (\ie, reinforcement approach, the copy-pointer mechanism) in abstractive summarization using the advancement in sequence to sequence model\cite{rush2015neural,nallapati2016abstractive,see2017get,henss2015reinforcement}. Although these approaches mainly focus on single-document summarization, There are other attempts at multi-document summarization \cite{yasunaga2017graph,cao2015ranking} as well.

Recent advancements in deep learning and transformer-based models \cite{li2019keep,liu2019topic} have achieved impressive performance in abstractive summarization tasks \cite{zhang2020pegasus,raffel2020exploring,lewis2020bart,zhu2020hierarchical}. Such transformer-based models are typically pre-trained on a large dataset and then fine-tuned on a smaller dataset to achieve impressive performance. There are also other methods to improve summarization using auxiliary tasks. Since summarization should contain all the salient information, It should generate answers to the logical question about the  input document.
Automatic question-answer (QA) generation in the process of summarization has shown promise in recent times \cite{guo2018soft,dong2020multi}. Such an automated QA generation method is used to verify if the generated summary entails the same information as the content by matching the answer generated from the content and the summary.

Text generation from multimodal data has always been a challenging research area in the NLG domain. Tasks like video captioning\cite{zhou2018end}, or summarization involve generating a compressed textual description of the data\cite{palaskar2019multimodal}. Recent developments show how these tasks can be benefitted from semantic representation learning in latent space that provides general-purpose embedding for downstream tasks \cite{lu2019vilbert,hubert2017learning}. Despite the performance, this approach limits due to controllability issues in tasks like summarization. As an alternative approach, there is also recent interest in utilizing Reranking-based approaches \cite{pernes2022improving} in abstractive summarization similar to machine translation  \cite{bhattacharyya2020energy}.

Evaluation of the summaries generated is a challenging task as there is no `single' correct summary for a dialogue \cite{lloret2018challenging}.  Numerous automatic metrics have been proposed for evaluating summaries  \cite{lin2004rouge,yogatama2015extractive,jung2019earlier,bert-score,hashimoto2019unifying,gao2020supert,sellam2020bleurt}. 
Human evaluation of summaries are another popular approach to evaluate the summaries, either by experts or by crowd-workers \cite{iskender2020towards,dang2006duc,khashabi2021genie}. 

\par
Our approach does not contribute to the development of a new model architecture for summarization instead it intends to benchmark and adapt the training methodology for incremental temporal summarization tasks. We adopt the current state-of-the-art transformer architecture and utilize transfer learning to generate summaries. We also evaluate the summaries ( generated by the experts) qualitatively using crowd-workers.

\cite{liu2022psp,tsimpoukelli2021multimodal,zeng2022socratic,pasca2023summarize}

\section{Task Formulation}

\subsection{Image frame sampling}
Since each video segment contains a lot of image frames based on its duration, it is essential to sample a fixed number of image frames for computational efficiency but to sample a fixed number of frames from a pool of frames that describe the entire event is tricky \cite{shi2019not}. We designed various experiments with and without sampling and observed that the middle frames of a video segment are the best frames that we can use to capture reasonable augmentation. For all the experiments we have performed, we used three frames from each video, \ie,  if N is the total number of image frames for a video, we use $N/2$, $N/2-1$, and $N/2+1$ frames.  \textbf{For ease of understanding we use "frames" to signify three middle frames for the rest of the discussion. } 

\par To know more about the details of the experiments we designed, Please refer to Appendix 1.
We have also designed a different network that learns to sample frames from the frame pool but we will keep this discussion for the sake of future directions of our work.

\subsection {Clustering-based batch creation}

In a single batch of data, instead of having similar event frames along with the corresponding event annotation, we performed a k-means-based clustering on the encoded feature of the image frames. Since similar event frames, in a single batch can not possess enough diversity, based on the clustering we can identify which event frame's features are dissimilar and use those features from different clusters to create a batch. Since frames is a collection of three middle frames we concatenate the features of each of these three middle frames and perform the clustering. This concatenation operation preserves the temporal relation between these middle frames for a particular annotation. This strategy improved our performance in augmentation generation compared to keeping similar event frames as depicted in the video data.

\begin{table}[h!]
\small
\centering
\resizebox{\columnwidth}{!}{
\begin{tabular}{|l| p{1.1cm}| p{1.1cm}| p{1.1cm}| p{1.1cm}|} 
\hline
\# samples &	TOP-1 (Kmeans) & TOP-1 (Random)& TOP-3 (Kmeans) & TOP-3 (Random) \\ \hline
150 (10) &	0.2156&	0.196&	0.549&	0.5098\\ \hline
300 (10) &	0.2749&	0.2745&	0.6176&	0.5098\\ \hline
1500 (10) &	0.2941&	0.2156&	0.6176&	0.598\\ \hline
150  (20) &	0.480392&	0.441176&	0.803922&	0.77451\\ \hline
300  (20) &	0.470588&	0.382353&	0.784314&	0.735294\\ \hline
1500 (20) &	0.303922&	0.245&	0.647059&	0.6372\\ \hline
\end{tabular} 
}
\caption{Clustering \& accuracy of semantic entities extracted K-means vs Random sampling.}
\end{table}

%\sumanta{need the kmeans table from the saples that matter slide}

\subsection {Perceiver Resampler}
Recent developments in transformer architectures \cite{jaegle2021perceiver} show we can scale transformers without the quadratic scaling concern in the attention. It involves learning a predefined number of latent input queries as input to the transformer and cross-attend to the feature. State-of-the-art vision-language-based model \cite{alayrac2022flamingo} architecture, utilizes this concept (perceiver resampler) to generate fixed-size embedding from variable length inputs.
\\
Traditional transformer-based image and text encoders use different kinds of pooling layers(mean/linear) to generate fixed embedding sizes from variable length input.
We replace the last pooling layer with the perceiver resampler architecture to get fixed-size output from both encoders in a similar fashion \cite{alayrac2022flamingo}, keeping the encoder layers frozen. This approach can also scale to larger inputs while retaining the model's expressivity.  As shown in the following table, using a learnable attention-based layer to generate fixed-size embedding compared to the pooling layer improves the feature quality. The Top1 accuracy for correctly predicting the annotation (semantic concept/augmentation) of the frame is more than other approaches.
\begin{table}[h!]
\resizebox{\columnwidth}{!}{
    \centering
    
    \small{
    \begin{tabular}{c p{2cm}}
       \hline
        architecture & accuracy (Top1\%/Top3\%)  \\ \hline
        pre-trained encoder & 18/27\\ %\hline
        pre-trained encoder+custom pooling & 21/32  \\
        \textbf{pre-trained encoder+perceiver resampler}& 26/38 \\  \hline

        \hline
    
    \end{tabular}
    }
 }
    \caption{ In custom pooling, we replace the pooling layer of the pre-trained model with a learnable pooling layer. for the learnable parameters, the result is for 5 epochs on the Youcook2 dataset. }
\end{table}

\section{Models}

We develop a two-stage approach (\rom{1}) \textbf{Phase I:} Learning the correct annotation from the video frame (frame to text). (\rom{2}) \textbf{Phase II:} Use these augmentations along with the summarizer's input to generate summarization (text to text) using pre-trained models (\ie, BART or distilBART).

\subsection{Phase I:}
We used the CLIP model to generate annotations from the video frames. CLIP models are known for learning visual concepts from language supervision. It involves two pre-trained encoders for image and text to predict the correlation between image and text. For image, CLIP uses a similar to ResNet50 architecture and for text, CLIP uses a masked self-attention Transformer. In order to train the CLIP model we used frames (as discussed in section 2.1) and corresponding annotation as input. Our experiments during feeding the data into the clip answer the following questions (\textbf{\rom {1}}) How to efficiently create a batch of data that consists of diverse examples?  (as discussed in section 2.2) (\textbf{\rom{2}}) Which frames in a pool of frame that describes a single event should be used for input? (as discussed in section 2.1).

\subsection{Phase II:}

It takes the semantic augmentation generated in phase I along with the transcription of the video as input to the pre-trained model and generates summarization (system flowchart in Figure 1.)

\par
Since each video is divided between segments based on the procedure, we can learn to predict the annotation (semantic concepts) for each of these segments from the video frames and use these concepts to augment the summarizer model's input, which is the transcript of the entire video, to generate summaries.

\section{Experimental Setup}

In Phase I, we finetuned the CLIP model for 1 epoch, and for the rest of the epochs, we only train the learnable perceiver resampler part keeping the encoder layers frozen. We observed finetuning the CLIP model for more than 2 epochs heavily degrades the performance on the prediction. Since it is already pre-trained on huge datasets, we found finetuning for one epoch on the new dataset is reasonable. For phase II also since we are using a pre-trained summarizer model, we adopt a similar strategy.

\subsection{Dataset}

We use Youcook2 for all our experiments. Existing datasets on instructional videos lack in many aspects (\ie, limited videos, limited actions etc.). The Youcook2 dataset is a collection of around 2000 cooking videos which contains around 89 cooking recipes and 14000 annotated clips with one descriptive sentence. 

\begin{table}[h!]
    \centering
    
    \small{
    \begin{tabular}{cc}
       \hline
        Dataset & Duration \\ \hline
        YouCook & 140 minutes\\ %\hline
        50Salads & 320 minutes  \\
        Breakfast& 34.25 hours  \\  %\hline
        \hline
        \textbf{Youcook2} & \textbf{176hours} \\
        \hline
    
    \end{tabular}
    }
 
    \caption{Comparison of other instructional video datasets}
\end{table}

Unlike Other datasets, as shown in Table 1, Youcook2 includes temporally localized procedure annotation with descriptions along with long-duration videos. Each video contains 3 to 16 procedure annotations. These procedure segments preserve rich semantic information which is useful for our task compared to other datasets. We randomly split the dataset to 67\% for training, 8\% for validation, and 25\% for testing.

%\subsection{Hyper-parameters}
%\sumanta{no idea...}

\subsection{Evaluation Metrics.} 
Our experiments are evaluated on the widely used evaluation metric Recall-Oriented Understudy for Gisting Evaluation (ROUGE score) for text-based summarization. It considers both the precision and recall between predicted and target summaries. Recall defines the proportion of words in the target summary generated by the predicted summary and precision defines the proportion of words generated by the predicted summary that appears in the target summary. ROUGE score has several methods and as shown in Table 1,
we evaluate on ROUGE-1(R-1)/ROUGE-2(R-2)/ROUGE-L(R-L) (Precision and Recall compare the similarity of uni-grams/bi-grams/Longest Common sub-sequence between target and generated summaries). For summarization, Recall is significant since it shows the generated summary captures all of the target summary's information. We gained a significant amount of improvement in recall using our method compared to the existing pre-trained model.

\begin{table*}[h!]
\small
\centering

\begin{tabular}{|l|l|l|l|l|} 
\hline
~~Experiments              & ~~\begin{tabular}[c]{@{}l@{}} Vision\\encoder\end{tabular} & ~\begin{tabular}[c]{@{}l@{}} 
~~~~~~Text\\~~~encoder\end{tabular}          & ~ ~\begin{tabular}[c]{@{}l@{}}~~~ 
~~~~~Batch mechanism\end{tabular}  & ~~\begin{tabular}[c]{@{}l@{}}~Result(\%)\\(Top1/Top3) \end{tabular}  \\ 
\hline
~~~~~~~~~~1             & ~ ~Yes                                                                   & ~ ~Yes          & k-means on image feature          & ~~~~39/70                                                              \\ 

\hline

~~~~~~~~~~2          & ~ ~Yes                                                                   & ~ ~Yes          & uniform sampling+k-means clustering          & ~~~~37/71                                                               \\ 

\hline
~~~~~~~~~~3    & ~ ~Yes                                                          & ~ ~Yes  & k-means on text feature & ~~~~36/68                                               \\ 
\hline
~~~~~~~~~~4 & ~ ~Yes                                                          & ~~ Yes(2nd last)  & Temporal k-means & ~~~~37/70 
\\

\hline
~~~~~~~~~~5 & ~ ~Yes                                                          & ~~ Yes& Temporal k-means & ~~~~41/73
\\
\hline
\end{tabular}
\caption{Results for CLIP-based semantic concept learning. Vision encoder and Text encoder column say whether it is frozen or not. For all the experiments, CLIP is finetuned for 1 epoch, and for the rest encoder layers are frozen }
\end{table*}

\begin{table*}[h!]
\small
\centering

\begin{tabular}{|l|l|l|l|l|} 
\hline
~ ~~Model              & ~ ~R-1 (P/R/F)          & ~ ~R-2 (P/R/F)         & ~ ~R-L (P/R/F)          & \begin{tabular}[c]{@{}l@{}}      ~~~Semantic\\Augmentation\end{tabular}  \\ 
\hline
BART-Base                                                                              & 0.49/0.42/0.44          & 0.26/0.23/0.23          & 0.46/0.41/0.42          & ~ ~ ~ ~NO                                                       \\ 

\hline

Distil-BART                                                                            & 0.50/0.50/0.48          & 0.26/0.27/0.26          & 0.46/0.46/0.44          & ~ ~ ~ ~NO                                                       \\ 

\hline
\textbf{BART-Base}                                                              & \textbf{0.61/0.53/0.54} & \textbf{0.36/0.33/0.33} & \textbf{0.58/0.51/0.52} & \textbf{~ ~ ~ YES}                                              \\ 
\hline
\textbf{Distil-BART}                                                           & \textbf{0.60/0.58/0.57} & \textbf{0.39/0.38/0.37} & \textbf{0.57/0.54/0.54} & \textbf{~ ~ ~ YES}                                              \\
\hline
\end{tabular}
\caption{P, R,F denotes precision-recall and F-1 score for the corresponding ROUGE scores.}
\end{table*}

\subsection{Results and Analysis}

Table 3 contains the results of experiments on CLIP-based semantic augmentation based on different batching of data. We found clustering based on the image features improves our performance rather than text features. We also uniformly sample the frames from the pool of frames as the input, which contains frames that are not contributing to the event depicted in the video (second experiment). We found adding more frames from the initial and end position of a video segment does not contribute to the accuracy much compared to middle frames. One of the experiments (experiment 4) includes training the text encoder's second last layer along with the perceiver resampler. we found the accuracy is less compared to a completely frozen encoder and learn only the perceiver resampler layer.

\par 
Table 4 contains the results of the summarization output.
we used the predicted semantic concepts along with the transcription as input to the pre-trained summarizer model. In order to show the efficiency of the augmentation, we did not fine-tune our pre-trained summarizer model with the semantic concept. Our approach shows significant improvement in terms of accuracy in all the metrics when we augment the input with predicted concepts from the CLIP model.

%\subsection {Qualitative comparison}

\section{Conclusion}
We presented two stage multimodal abstractive video to text summarization model that takes advantage of the extra semantic concepts along with summarizer input. We have provided detailed evaluation for each of our step. We demonstrate that our method gains a significant improvement over the existing pre trained summarizer model. We use this semantic augmentation generation step as an intermediate process. We also showed using various methods like adding a perceiver resampler layer and batching using k-means based clustering with temporal relation can improve the accuracy for concept generation which in turn improves the summarizer quality.

\bibliography{anthology,custom}
\bibliographystyle{acl_natbib}

\appendix

\section{Example Appendix}
\label{sec:appendix}

This is an appendix.

\end{document}